# Improving Retrieval-Based Question Answering with Deep Inference Models

George-Sebastian Pîrtoacă[1], Traian Rebedea[1], Ștefan Rușeți[1]

[1] University Politehnica of Bucharest, Faculty of Automatic Control and Computers
george.pirtoaca@stud.acs.upb.ro, {traian.rebedea, stefan.ruseti}@cs.pub.ro

*Abstract* - **Question answering is one of the most important and difficult applications at the border of information retrieval and natural language processing, especially when we talk about complex questions which require some form of inference to determine the correct answer. In this paper, we present a two-step method that combines information retrieval techniques optimized for question answering with deep learning models for natural language inference in order to tackle the multiple-choice question answering problem. In the first stage, each question-answer pair is fed into an information retrieval engine to find relevant candidate contexts that serve as the underlying knowledge for the inference models. In the second stage, deep learning architectures are used to predict if a candidate answer can be inferred from the context extracted in the first stage. We deploy multiple deep learning architectures pre-trained on different datasets in order to capture semantic features and to enlarge the scope of the questions we can answer correctly. As it will be described, each dataset used for training the inference models has particular characteristics that can be exploited. In the end, all these solvers are combined in an ensemble model to predict the correct answer. This proposed two-step model outperforms the best retrieval-based solver by over 3% in absolute accuracy. Moreover, the model can answer both simple, factoid questions and more complex questions that require reasoning or inference.**

*Keywords - question answering, neural networks, deep learning, natural language inference, information retrieval.*

## I. Introduction

We live in a world where Artificial Intelligence is becoming more and more a part of everyday life. The ability to answer questions expressed in natural language is very important in order to deploy successful products requiring natural language interaction. An increasing number of commercial systems that handle computer-human interaction encompass a form of question answering (QA). Thus, Google Assistant and Amazon Alexa manage to answer simple factoid questions (such as "What is the most viewed song on YouTube in the current month?"). All these systems have a common trait as they are optimized to answer only a specific category of factoid questions and they rely on extracting information from a large collection of sources (e.g. proprietary knowledge bases, large collection of web pages). However, question-answering systems should be able to answer more complex questions, that are not only retrieval-based but also require inference over the extracted information. Examples of such multiple-choice questions are related to middle school science classes (e.g. "Which of the following has been discovered first? A) car B) phone C) airplane D) wireless communication"). For this question, it is very unlikely that the extracted knowledge (e.g. from Wikipedia) contains exactly the correct answer to the question. The more realistic scenario is that the extracted contexts for each candidate answer also mention the date of invention for each technology. Then, a QA system should compare these dates to infer which is the oldest among the four choices.

The premise is that these questions are more difficult to answer (even if the system is given a list of possible candidate answers) and require models that should be able to (partially) understand the context in which the question is asked and to perform some kind of inference to determine the correct answer. The datasets used to train and evaluate the models proposed in this paper contain both factoid (e.g. „*What is the population of South Africa?*") and non-factoid questions. Schoenick et al. [2] have argued that answering science questions with high accuracy can be a good way to assess machine intelligence. Given these observations, the problem approached in this paper is answering multiple-choice questions from a field of science studied in grade school (e.g. chemistry, biology, physics, and astronomy). All questions are expressed solely in natural language (without diagrams or equations) and each question has four candidate answers, out of which exactly one is correct. The system can use any publicly available resources (e.g. Wikipedia, science books, ad-hoc data corpora) as a knowledge base. These resources are used to extract the knowledge (mostly domain-specific) required to answer the questions. The input of the system is the question itself along with the four candidate answers, while the output is the predicted answer together with a confidence score. In this paper, we propose to improve standard retrieval based techniques used for question answering with deep learning methods [3] which have been shown to provide good results for assessing (simple) natural language inference. Deep learning models are top performers on various reading comprehension and natural language inference datasets, such as SQuAD [16], SciTail [21], or MultiNLI [23]. Moreover, deep learning has greatly improved machine translation performance, which is also a problem that requires a form of reading comprehension.

The paper continues with an overview of the domain, focusing on the solutions proposed for solving the multiple-choice science question answering problem. Then, we present our proposed method in detail and evaluate the model's performance on relevant datasets.

## II. RELATED WORK

A lot of work has been invested to improve retrieval-based baselines for question answering (QA) and to give a better sense of what a QA system could achieve with proper, structured knowledge and more complex ranking strategies for the candidate answers. We can split the strategies used for tackling the question answering problem into two big categories: using structured information with symbolic reasoning and using non-symbolic reasoners such as neural networks or SVMs. The latter require larger data volumes for training but were shown to generalize better than symbolic methods and currently they are the top performers on various reading comprehension datasets such as SQuAD [16], SciTail [21], or MultiNLI [23].

### A. Symbolic approaches

Khot et al. [4] propose a way to formulate the multiple-choice question answering problem as a Markov Logic Network (MLN) problem. Each question along with its four candidate answers is translated into four questions that can be answered with either *"True"* or *"False"*, thus transforming the QA problem into a binary classification. Then, the system has to decide what questions can be answered with *"True"*. The correct answer is the one with the greatest confidence (probability) as predicted by the system. A knowledge base is used, consisting of rules automatically extracted from natural language corpora and represented as IF-THEN clauses, as described by Clark et al. [5]. Three MLN formulations have been proposed and evaluated: First-order MLN, Entity Resolution Based MLN, and Probabilistic Alignment and Inference (PRALINE).

Khashabi et al. [6] describe a method called TableILP that treats the multiple-choice QA as a sub-graph optimization problem. It uses semi-structured information represented in tables where each row is a predicate of arity *k* (number of columns) over a short natural language sentence. The QA problem is viewed as an optimal sub-graph selection problem, where the algorithm tries to find the *(question, answer)* pair that best fits the knowledge base.

$KG^2$ [1] is a model that employs contextual knowledge graphs in order to find the correct answer to a question. Using a large text corpus as the background knowledge, information is extracted, and relation triples are constructed from each sentence using Open IE (Manning et al. [26]). These relations are combined into a contextual knowledge graph that is further used for reasoning. The question and the answer are combined into a hypothesis that is also transformed into a knowledge graph. Thus, the entailment problem is transformed into a graph ranking problem: a graph scoring function f: G x G → **R** is leaned so as to assign high scores to pairs of graphs that represent correct answers to the given questions.

### B. Non-symbolic approaches

Jansen et al. [7] use potential answer justifications for each candidate answer not only to improve the system's accuracy but also to provide simple, natural language explanations for the predicted answer. These are useful when one needs high confidence in the answer provided by the system (for example, in the medical domain). Information is extracted from various sources (e.g. study guides) and decomposed into smaller sentences (called information nuggets). Combining those information nuggets results in potential answer justifications. A perceptron is trained to order the list of justifications and to choose the most reliable one. The computed answer is the one that corresponds to the best justification.

Nicula et al. [8] proposed a model for predicting correct answers based on candidate contexts extracted from Wikipedia. Using Lucene-based indexing and retrieval, each paragraph in the English Wikipedia is indexed and used as a candidate context for questions and corresponding answers. Each *(question, answer)* pair is searched in the index and the top 5 retrieved documents, along with the question and candidate answer, are concatenated and fed into a deep neural network that computes a score for the *(question, candidate, context)* triple. Two neural network architectures are tested, that use different ways of combining the question, candidate, and context. For each architecture, different encoders have been evaluated: bidirectional long-short term memory network (BiLSTMs) [9] and convolutional neural networks [10, 11].

The most similar approach to the one presented in this paper has been described by Clark et al. [12]. The proposed model combines an information retrieval solver, statistical information using Pointwise Mutual Information (PMI), text similarity using word embeddings to represent the lexical semantics and a simple Support Vector Machine ranker, and a structured knowledge solver. All of these solvers have been combined in a subsequent step using a logistic regression classifier. The method has been tested on the NY Regents 4th Grade Science exams and it outperforms any other solutions on that dataset.

Our proposed method is different from the described approaches in the way it combines several solvers that work at different representation levels: information retrieval and natural language inference using deep learning. Following this strategy, our model is able to answer both easy, factoid questions and more complex questions that require textual inference.

## III. PROPOSED METHOD

We propose a two-stage model that combines an information retrieval (IR) engine with several deep learning architectures (called *solvers*) in order to compute the correct answer. The purpose of the first stage is to extract relevant knowledge support (called *contexts*) that is closely related to the question and the candidate answers. As plain knowledge sources, we index the entire English Wikipedia, the ARC Corpus[1], as well as science books available on CK12[2]. In the second stage, different neural networks are fed with a *(question, answer, context)* triplet and trained to predict the likelihood that the answer is correct for the question at hand given the context extracted by the IR engine as additional knowledge. We used multiple neural networks pre-trained on different natural language processing tasks (as is described in the next sections) and then fine-tuned for multiple-choice question answering. Each model in the ensemble computes the likelihood for each triplet it receives. These scores are further fed into a simple neural network that computes the final score as a combination of the scores of each solver (voting mechanism). The final predicted answer is the one with the highest score as computed by the voting mechanism.

---

[1] http://data.allenai.org/arc/arc-corpus/ (last access, December 2018).
[2] https://www.ck12.org/ (last access, December 2018).

Thus, we have divided the main QA task into two sub-tasks, which can be designed and tested independently:

1. Extract relevant contexts for each *(question, candidate answer)* pair using an information retrieval approach.

2. Construct various (more complex) models to predict if an answer is correct for the given question based on additional information which can be inferred from the contexts extracted in the previous step.

We postulate that using multiple solvers in an ensemble model makes the QA system capable of finding the correct answer to more complex questions. This hypothesis is supported by the fact that the dataset used to assess the performance of the proposed model contains various questions from different science domains (e.g. chemistry, biology), which require different reasoning techniques to infer the correct answer. Another aspect that supports our hypothesis is that we use different datasets in order to pre-train the neural networks, datasets which are essentially different in their core characteristics (see the pre-training section for more intuition).

In our research, we decided to use deep neural networks for the second stage solvers, as they have been recently shown to accurately solve (simple) textual inference problems [21]. The ensemble stage uses another neural network for combining the scores computed by each solver. However, any other non-linear model can be used as the voting mechanism. Neural networks have been preferred here because they are expressive enough to learn complex hypothesis and also because they are easier to work with especially in the framework that we deployed for implementing our QA system.

In the following sections, we describe each component of the proposed solution, focusing on the solvers in the second stage of the model. In the end, we evaluate both the system as a whole and the performance of each solver in particular.

A. *Extracting relevant knowledge support*

As mentioned, the first step requires to generate relevant contexts for each candidate answer and the corresponding question. This is basically a retrieval problem over a corpus of documents. We use Lucene[3] to index a large collection of documents (entire English Wikipedia, science books collected from CK-12, and ARC Corpus) pertinent for the science QA task at hand and later to retrieve relevant information for candidate answers. The raw text from each source (e.g.

---

[3] https://lucene.apache.org/core/ (last access, December 2018).

Wikipedia) is split into shorter text-spans (maximum of 10 consecutive sentences) with some overlap between them. That is, the first two sentences of a text-span are also the last two sentences of the previous paragraph. The overall stride has been chosen empirically. In total, we indexed about 15 GB of raw text, most of which was generated from the Wikipedia's March 2018 dump (>85%). However, the science book collection and ARC Corpus turned out to be very useful in increasing the accuracy of the system. The reason behind this behavior is that our QA dataset is science focused and both corpora contain information very concentrated and aimed at the science field, whereas the Wikipedia corpus covers a wider spectrum of information that is not always relevant to science questions.

To obtain the contexts, we run a Lucene query over the indexed documents with the following query string: "*<question>* AND *<candidate answer>*". Lucene ranks the results according to the TF-IDF score [13] between the query string and each document in the index. We experimented with collecting top *N* documents returned by Lucene (*N=1..5*), but there was no clear advantage in performance. Thus, we ended up using only the first ranked document as context for a candidate answer.

The IR engine serves two purposes. First, it provides the relevant context that is required to perform textual inference in the next step. Second, it defines the first solver of the ensemble – the TF-IDF scores as computed by Lucene are fed into the final voting mechanism. Locally, the answer with the highest TF-IDF score is the most probable to be picked as the correct one, but this decision can be changed after looking at output scores from the second-stage inference models. The purpose of the voting mechanism is to learn how to effectively combine all these scores to maximize the end-to-end accuracy. Moreover, the TF-IDF score without the voting mechanism offers a simple baseline for testing the accuracy of the system: given a question and all its candidate answers, pick the answer with the largest score retrieved by Lucene as the predicted answer. We are going to refer to this method as the *IR Solver* – the predicted answer is the one with maximum TF-IDF score. This baseline is completely unsupervised and uses no textual inference or more linguistic processing. Thus, we can easily compare how adding additional solvers – trained to perform different types of natural language inference – can increase the performance of the model.

One important observation is that when querying the Lucene index for fetching contexts, we use a term-based weighting in computing the TF-IDF scores. This means that for each term in the question we introduce an importance score which acts as an additional term weight in the TF-IDF score computation. The importance score depends on the semantic essentialness of the term. A term is essential in a question if its removal makes the question impossible to answer (e.g. "ocean" is essential in "What is the largest ocean on Earth?"). The final TF-IDF score is influenced more by terms which are considered relevant for understanding the question. In order to learn which terms are important and which are not, we used the technique described in [24]. A neural network classifier based on semantic and syntactic features (such as pointwise mutual information or part-of-speech tags) is trained using a labeled dataset collected by Khashabi et al. [14] consisting of 2.2K questions annotated with importance scores for each term. The authors of the model [24] showed that term importance scores improve the quality of the returned contexts as they are capable of distinguishing noisy terms from critical terms that are essential in understanding the question at hand. This can be observed in Fig. 1, which presents the term importance scores computed by the model for a sample question.

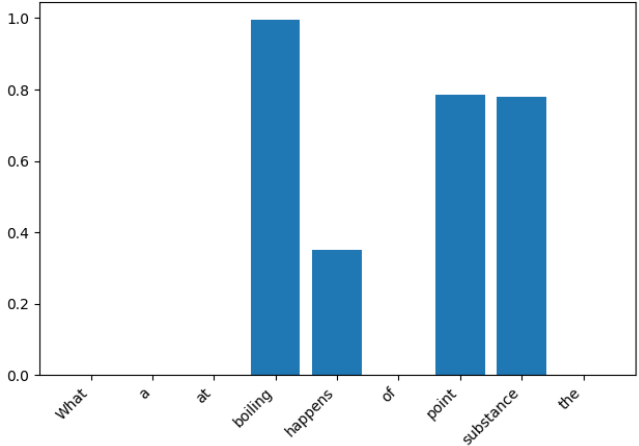

Fig. 1. Importance scores for the question "What happens at the boiling point of a substance?"

Having computed a meaningful candidate context for each *(question, answer)* pair, more complex natural language inference, and semantic similarity computation must be performed to find the best answer (apart from the TF-IDF score which is also used as input in the final voting mechanism). For this, two different strategies are deployed by the second stage solvers as is further detailed.

### B. Improved semantic similarity computation

The first improved linguistic processing is to compute a more efficient semantic similarity using word embeddings and recurrent neural models for sentence representation. Thus, we train a neural network to predict whether the tuple *(question, answer, context)* is the most suitable to contain the correct answer using an improved semantic representation for the given triple. As the science QA datasets are rather small, we use transfer learning by training the model on larger datasets (as described in the following sections).

This solver takes a tuple *(question, answer, context)* as input and predicts whether it can infer the current answer is the correct one. For this task, we adapt the Bidirectional attention flow (BiDAF) architecture proposed by Seo et al. [15] for reading comprehension, a special case of QA. We use the same character and word embedding layers as input layers. Bidirectional attention flows from question to context (and vice versa). The answer itself is encoded using both a Char-CNN and pre-trained GloVe embeddings [17], and then passed through a single layer of a bidirectional LSTM where it is summarized into a single dense vector. We remove the last dense layers of the original BiDAF architecture (that follow the modeling layer) and instead concatenate the BiDAF encoding of the question and context with the answer encoding (on the time axis). After that, the model transforms the *(question, context)* encoding and the answer representation via a bidirectional LSTM. The output stage of the network is a feed-forward architecture with 3 fully connected layers and dropout layers with 0.20 probability to reduce the overfitting problem. The last dense layer is a softmax with two classes (0 – wrong answer, 1 – correct answer). The complete architecture is represented in Fig. 2. This adapted BiDAF architecture will be referenced in the remainder of the paper as the *QA-BiDAF*.

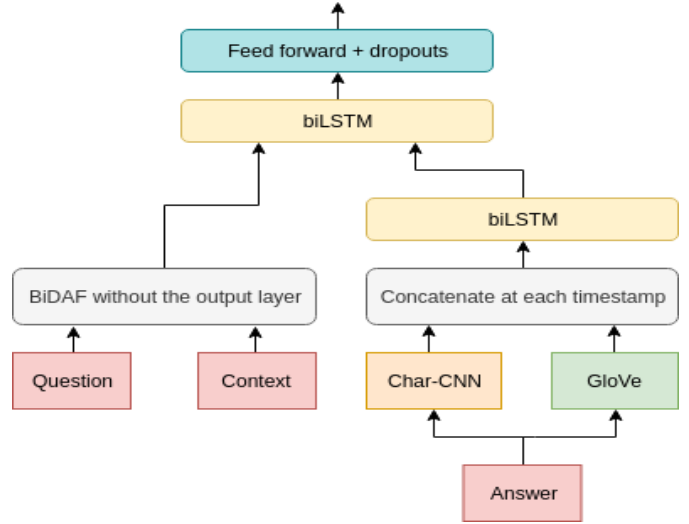

Fig. 2. BiDAF architecture for multiple-choice QA (QA-BIDAF)

The character embeddings are obtained from GloVe word embeddings by taking the average for all words that contain the character. The embedding vectors are further reduced to 35 dimensions using PCA. The BiDAF model proposes several important improvements over vanilla LSTM models when exposed to large input sequences with long dependencies [15]. In general, the contexts extracted with the information retrieval engine are rather large (~300 words, on average). However, only some shorter text spans can be used to actually predict the correct answer. Consider the following scenario, in which the context returned by the IR component is the following: *"The general theory of relativity which was developed by Einstein between 1907 and 1915 is one of the most important theories in physics. ... [Theory description here]... It explains the inner workings of the gravitational force and how it is linked to other forces in nature."*. This context is generated for the question: *"Between what years did Einstein develop the theory which explains the source of the gravitational field?"*. In this case, it is natural to skip the part of the context which provides details about the general theory of relativity as it does not offer any information helpful in answering the question at hand.

Furthermore, some science questions are intentionally injected with extra information that is not meaningful in finding the correct answer. This is actually a pedagogical objective as teachers want to assess whether students are able to detect the relevant information in a question. The QA-BiDAF model can learn to skip over parts of the question when computing the overall text representation.

In our implementation of the QA-BiDAF model, we used pre-trained GloVe 50D word embeddings [17]. One may argue that GloVe captures similarities between words from the same semantic field, but with slightly different interpretations (e.g. "man" and "woman", "queen" and "king"). This issue can be crucial for questions answering, where such subtle differences in meaning are important for determining the correct answer. This is one of the reasons for which the BiDAF model also uses char-level embeddings [11] and a highway layer [18] capable of fine-tuning specific word embeddings.

The hidden size of the bidirectional LSTMs is 100 and the fully connected layers have 100 neurons with ReLU activations (except for the last layer which has only two neurons and a softmax activation is applied). We employ dropout with 0.20 probability after each biLSTM and dense layer to avoid overfitting. Inside the LSTM cells, a recurrent dropout is applied with 0.15 probability.

TABLE I. QUESTION TO STATEMENT TRANSLATION EXAMPLES

| Original question | Statement |
|---|---|
| Which of these is a greenhouse gas? | @placeholder is a greenhouse gas. |
| What do plant roots prevent? | Plant roots prevent @placeholder. |
| What does FIFA stand for? | FIFA stands for @placeholder. |
| Where is corruption most noticeable? | Corruption is most noticeable in @placeholder. |
| Who is the Microsoft owner? | @placeholder is the Microsoft owner. |
| When was ENIAC fully operational? | ENIAC was fully operational on @placeholder. |
| What year did Chopin die? | @placeholder (year) Chopin died. |
| Which ocean does Portugal border? | @placeholder (ocean) Portugal border. |

*C. Multiple-choice QA as natural language inference*

The second method to improve the IR baseline for the multiple-choice question answering task is to use natural language inference (or entailment). To achieve this, first, we transform the pair *(question, answer)* into an affirmative sentence that forms a hypothesis. For example, the question *"Which is the most successful team in the NFL?"* combined with the candidate answer *"LA Rams"* produces the following hypothesis: *"LA Rams is the most successful team in the NFL"*. In this setting, the context extracted by the IR engine tuple will act as the premise. Using this simple transformation, the multiple-choice QA task has been reframed into textual entailment problem.

First, the question is translated into an affirmative statement containing a special token @*placeholder* where the answer needs to be filled in. Then, we replace @*placeholder* with the candidate answer to determine the hypothesis. Using an empiric approach, by investigating the questions in several multiple-choice QA datasets, we have manually identified 36 different question types (for example, "what" questions, "why" questions, etc.). For each question type, a set of rules (mostly based on the syntactic dependency trees generated using spaCy[4]) has been proposed in order to translate from the question to the corresponding affirmative statement. A rule is basically transforming a subtree of the dependency tree into another subtree. On average, each question type has four rules and an associated priority for each rule. Therefore, a set of around 150 rules is used for this task. Those rules have been identified in a purely manual process. We have made the source code that performs this transformation public to be used for similar situations[5]. Please refer to Table 1 for a list of translation examples. Having a *(question, answer)* pair translated into an affirmative sentence, we deploy the same BiDAF architecture to perform natural language inference. We modified the output layer to fit our needs by replacing it with a 3-way softmax layer: entailment, neutral, or contradiction. This version of BiDAF will be referred to as the *NLI-BiDAF*. The hyperparameters of the architecture are the same as the ones described for the QA-BiDAF architecture, the main difference being that the model now has only two inputs – the premise and the hypothesis instead of three – the question, the answer, and the context.

*D. Combining all solvers*

We have described the three main approaches proposed in this paper to solve the multiple-choice QA problem: using information retrieval, using enhanced semantic similarity computation for question answering, and using natural language inference. In the last stage, we combine all these models into a single one using another fully connected network (2 hidden layers with 5 and 20 neurons followed by a single sigmoid neuron to output the final probability) taking the role of an ensemble. To summarize, for a *(question, answer, context)* tuple, multiple predictors in the second stage of the system give the following results:
 a. The TF-IDF score from Lucene;
 b. The QA score from the QA-BiDAF;
 c. The NLI-BiDAF scores from the natural language inference model.

The output of the voting network is the final likelihood of an answer being correct. In the end, the candidate answer with the highest score is selected as the predicted answer.

IV. PRE-TRAINING MODELS

We have described some neural networks architectures that are going to be trained on the available multi-choice QA datasets. One major shortcoming that kept away deep learning models from being used in the multiple-choice science QA task is the limited amount of data. Training a neural network such as BiDAF requires a significant dataset due to its large number of parameters (in our implementation, 1.2M trainable weights). Unfortunately, all the multiple-choice science QA datasets available at this moment (that are publicly available to the best of our knowledge) contain too few examples to successfully train large, deep neural architectures (less than 10K examples in datasets). On the other hand, open-ended QA datasets like SQuAD [16] are large enough to allow complex

---

[4] https://spacy.io/ (last access, December 2018)

[5] https://github.com/SebiSebi/AI2-Reasoning-Challenge-ARC (last access, March 2019)

neural networks to be trained. We would like to use datasets like SQuAD to pre-train our models (QA-BiDAF and NLI-BiDAF). The only problem is that SQuAD contains open questions, not multiple-choice ones. Thus, in order to employ transfer learning, we need to transform the SQuAD dataset into a multiple-choice QA dataset. This can be done to some extent just by looking at how SQuAD has been constructed. One important aspect of the SQuAD dataset is that for each question a relevant context is provided which also contains the correct answer. Furthermore, there are multiple questions addressing the same context. We use this property to generate wrong candidate answers and ultimately to transform the SQuAD dataset into one suitable for multiple-choice QA. The method we propose is to use the fact that for any context, correct answers for a question can be considered as wrong candidate answers for the other questions referring to the same context. Therefore, in order to generate wrong candidate answers to a given question, we look at all other questions based on the same document and randomly pick correct answers for these questions. This assures that, to some extent, wrong answers to a question, are not trivially "wrong" (e.g. on a completely different topic). For example, we may have the following two questions given the same article: *"Which NFL team represented the AFC at Super Bowl 50?"* and *"Which NFL team represented the NFC at Super Bowl 50?"*. If we look at all the possible correct answers to the second question these are clearly wrong answers for the first one. Apart from that, a system which pretends to answer both questions needs to have some comprehension ability in order to distinguish between the candidate answers. In order to pre-train the neural networks (QA-BiDAF and NLI-BiDAF), we used four large datasets for reading comprehension and natural language inference, each with its special characteristics:

    a. The modified SQuAD v1.1 dataset adapted for multiple-choice QA as described in this chapter. This dataset is used to train the QA-BiDAF whereas all the other datasets are used for pre-training the NLI-BiDAF model.

    b. The Stanford Natural Language Inference (SNLI) corpus [19] – a collection of 550,000+ English sentence pairs manually labeled with entailment, contradiction, or neutral. All sentences are based on image captions and are limited to descriptions of visual scenes.

    c. The Multi-Genre NLI (MultiNLI) corpus [20] is similar to the SNLI corpus but offers a wider range of genres (not only visual, image captions). The dataset contains over 415,000 English sentence pairs from fiction, travel, and government sources.

    d. The SciTail dataset [21] consists of 27,000 entailment pairs of sentences (like SNLI or MultiNLI). This dataset is particularly important since it contains only science related sentences.

The models resulted from training the adapted BiDAF architecture on these four particular datasets are independent solvers due to the core differences in the dataset structure, genre, and semantic content. The hypothesis is that the neural networks have learned to encode core characteristics that are needed in general reasoning (semantic similarity, textual inference) and these can be transferred to answer new, unseen questions even of a slightly different nature (domain, method of generation). The parameters of all the models are later on fine-tuned for the multiple-choice QA downstream task on specific datasets as described in the next section. Those datasets are small in size, containing 1K or 2K questions for training and for testing depending on the version of the dataset (ARC Dataset). Both QA-BiDAF and NLI-BiDAF have been trained using the Adam optimizer [27], with batch size 64, decreasing the multi-class cross-entropy loss.

To sum up, the proposed QA system can be summarized as follows: each pair *(question, candidate answer)* is searched in a large collection of documents (Wikipedia, science books, ARC Corpus), then multiple inference models pre-trained on large datasets are used to predict if the answer is correct given as additional information the most suitable context extracted from the documents. Both the TF-IDF score (*IR Solver*) from the first stage and the 4 different scores (*QA-BiDAF* and *NLI-BiDAF* for SNLI, MultiNLI, and SciTail) in the second stage are combined using a voting mechanism.

V. RESULTS

We evaluate our model on the AI2 Reasoning Challenge (ARC) dataset proposed by Clark et al. [22], which contains 7,787 grade-school level science-related questions, partitioned into a challenge set and an easy set, based on the difficulty of the question. The easy dataset contains mostly factoid questions that do not require complex inference procedures. The challenge dataset contains more complex questions, which both a retrieval and a co-occurrence method fail to answer correctly [22]. In Table 2 and Table 3, *IR Solver* refers to the information retrieval model, where the predicted answer is the one with the highest TF-IDF value (stage two models are not used at all). By comparing to the IR Solver, our proposed two-stage model increases the accuracy by 3.08% on the Challenge test dataset and by 8.47% on the Challenge Dev dataset. However, on the Easy test dataset, the performance is not boosted by the two-stage model due to the nature of the questions and the fact that IR outperforms all inference solvers. Questions in the ARC Easy dataset are mostly factoid and they require a good retrieval engine rather than more complex inference techniques. It is enough to extract relevant contexts and the answer will appear inside the context as a text span. This can be easily seen in the following example: *"Earth's core is primarily composed of which of the following materials? a) basalt b) iron c) magma d) quartz."*

It is important to notice that on the Challenge dataset the best inference model (BiDAF trained on MultiNLI), outperforms the IR Solver on its own. One possible explanation for the fact that the model pre-trained on the MultiNLI dataset performs the best on challenge questions is that the MultiNLI is a multi-genre, multi-domain dataset, rendering many important phenomena such as visual scenes and temporal reasoning (today, yesterday). It is also large enough to allow pre-training and information transfer. Furthermore, the accuracy of the two-stage model is 1.19% higher than the BiDAF (MultiNLI) model, suggesting that the voting mechanism has learned to combine the scores in such a way that the end-to-end performance is increased even though the local components individually perform worse. It is also useful to take a look at some of the questions that are answered incorrectly by the information retrieval model but are answered correctly by the proposed two-stage model:

- „*All organisms depend on the transfer of energy to survive. Which best shows the energy transfer between animals in a shoreline ecosystem?*" A) Fish -> Plants -> Birds; B) Plants -> Birds -> Fish; C) Plants -> Fish -> Birds; D) Fish -> Birds -> Plants;
- „*Patricia and her classmates are visiting different rocky areas in the city. They want to know what kind of rocks can be found in each area. If the investigation is done correctly, what will Patricia and her classmates do each time they visit an area?*" A) draw the rock shapes; B) count the rock numbers; C) record the rock types and locations; D) measure the rock masses and lengths;
- „*Dr. Wagner is investigating a newly discovered, disease-causing agent. She determines that one structure in the agent is double-stranded RNA. What kind of agent is Dr. Wagner studying?*" A) a virus; B) a protis; C) a fungus ; D) a bacterium;

All these questions require some form of simple textual inference to determine the correct answer (information retrieval is not enough). In the second example above, the system has to reason about the fact that by recording the type of rocks and the locations where they are found, Patricia and her classmates can further find which rocks can be found in what areas. Our hypothesis is that the inference solvers are trained to perform exactly this kind of logical deduction and thus, improve the accuracy compared to the IR solver. It is important to note that the 3% absolute improvement means more than 10% relative improvement compared to the IR Solver.

In comparison to other models – see ARC Leaderboard[6], our combined solution ranks second on the Easy dataset and eighth on the Challenge dataset. Moreover, at the moment of writing this paper, our model is the only one that performs well on both Easy and Challenge datasets. This is because it combines solvers that work at different levels and tackles the questions from various directions: simple information retrieval, more complex textual inference, and enhanced semantic similarity. Table 4 offers a comparison with other solutions that solve both Easy and Challenge questions. By performing an error analysis on a subset of the ARC Challenge dataset, we observed that ~50% of the questions are answered incorrectly due to insufficient support from the knowledge base and the IR component. This means that for about half of the questions answered incorrectly, the extracted contexts are not helpful in finding the correct answer (e.g. even a human would not be able to pick the correct answer given only the information in the context). Therefore, using the proposed two-stage strategy, the second stage solvers can only improve the results for about half of the questions in the dataset.

---

[6] https://leaderboard.allenai.org/arc/submissions/public (last access, December 2018)

TABLE II.  P@1 OF THE TWO-STAGE COMBINED MODEL AND INDIVIDUAL SOLVERS ON THE ARC EASY DATASET

| Model | ARC Easy Dev | ARC Easy Test |
|---|---|---|
| Random choice | 25.00% | 25.00% |
| QA-BiDAF (SQuAD) | 25.40% | 26.89% |
| NLI-BiDAF (SNLI) | 27.34% | 29.94% |
| NLI-BiDAF (MultiNLI) | 29.98% | 32.18% |
| NLI-BiDAF (SciTail) | 30.16% | 30.57% |
| IR Solver | 61.38% | 61.10% |
| **Two-stage model** | **61.90%** | **61.10%** |
| **Difference** | **+0.52%** | **+0.0%** |

TABLE III.  P@1 OF THE TWO-STAGE COMBINED MODEL AND INDIVIDUAL SOLVERS ON THE ARC CHALLENGE DATASET

| Model | ARC Challenge Dev | ARC Challenge Test |
|---|---|---|
| Random choice | 25.00% | 25.00% |
| QA-BiDAF (SQuAD) | 24.75% | 22.15% |
| NLI- BiDAF (SNLI) | 27.12% | 24.64% |
| NLI- BiDAF (MultiNLI) | 25.76% | 25.67% |
| NLI- BiDAF (SciTail) | 30.51% | 23.00% |
| IR Solver | 22.71% | 23.78% |
| **Two-stage model** | **31.18%** | **26.86%** |
| **Difference** | **+8.47%** | **+3.08%** |

TABLE IV.  OUR PROPOSED MODEL COMPARED TO OTHER SYSTEMS

| Model | ARC Easy Test | ARC Challenge Test |
|---|---|---|
| TableILP ([6]) | 36.15% | 26.97% |
| KG$^2$ [1] | not reported | 31.70% |
| Nicula et al. [8] | 30.30% | 26.33% |
| Decomposable Attention [22, 25] | 58.27% | 24.34% |
| **Two-stage model (ours)** | **61.10%** | **26.86%** |

One important advantage of the proposed two-stage model is that it brings together different natural language solvers that reason about the contexts generated by the retrieval-based models. The overall performance can be improved in both stages: better retrieval methods can generate more relevant contexts, while adding more solvers working on different levels can improve the second stage inference processes. The current model highlights that retrieval-based QA can be easily improved by adding deep learning solvers for more advanced natural language processing (inference, semantic similarity) without adding any complex feature engineering, rules, or linguistic expertise. Moreover, this is one of the first approaches that shows transfer learning can be useful for improving QA systems on domains with small datasets.

## VI. Conclusions

In this paper, we described a two-stage model that combines different solvers to tackle the multiple-choice science question answering problem. In the first stage, we deploy an information retrieval engine working on different knowledge bases (Wikipedia, science book collection, ARC Corpus) to extract relevant candidate contexts for each *(question, candidate answer)* pair. In the second stage, we employ more complex models based on deep neural networks to further analyze each *(question, candidate answer, context)* triplet to determine whether there is a natural language inference relation or a more complex semantic similarity between the context, question, and candidate answer. In the end, all solvers, including the IR one (TF-IDF scores), are combined using a voting mechanism implemented as a neural network to determine the most probable answer.

The models have been optimized using continued training to overcome the rather small datasets available for multiple-choice science QA. Our system outperforms the retrieval-based models on more challenging questions, where an information retrieval solver alone is not powerful enough to determine the correct answer. Attention mechanisms allow deep neural networks to focus only on the important aspects of the extracted candidate contexts and ignore large, irrelevant spans.

Some ideas introduced in this paper, e.g. multiple-choice SQuAD or transforming questions to affirmative statements, can be used in other applications as well. We also showed that transfer learning is useful for improving QA systems on specific domains (e.g. science) with small datasets. The proposed model can be improved by using a better knowledge base to find candidate contexts and by adding additional solvers to the ones proposed in this paper. For example, solvers that use structured information to tackle even more difficult questions. Moreover, our model would benefit from larger datasets (science specific) to train the neural networks without the need for pre-training. Thus, the models would be able to learn more complex hypothesis for textual inference related to science questions, increasing the accuracy of the QA system.